%% file: bare_jrnl.tex
\makeatletter
\def\eg{\emph{e.g}} 
\def\ie{\emph{i.e}}

\def\etal{\emph{et al}}
\makeatother

\documentclass[journal]{IEEEtran}
\usepackage{graphicx}
\usepackage{cite}
\usepackage{amsmath}
\usepackage{amssymb}

\usepackage{booktabs}
\usepackage{makecell}
\usepackage{multirow,xspace}
\usepackage{amsmath, eucal}
\usepackage{caption}
\usepackage{subcaption}
\usepackage{xcolor}
\usepackage{bm}
\usepackage{url}
\newcolumntype{I}{!{\vrule width 1pt}}

\usepackage{array}

\begin{document}
\title{Calibrating Class Activation Maps for Long-Tailed Visual Recognition}

\author{Chi Zhang,
        Guosheng Lin\thanks{Corresponding author: Guosheng Lin.},
        Lvlong Lai,
        Henghui Ding,
        Qingyao Wu
          % ,~\IEEEmembership{Life~Fellow,~IEEE}% <-this % stops a space
\IEEEcompsocitemizethanks{\IEEEcompsocthanksitem  Chi Zhang, Guosheng Lin, and Henghui Ding are with School of Computer Science and Engineering, Nanyang Technological University (NTU), Singapore 639798 (email:  chi007@e.ntu.edu.sg, gslin@ntu.edu.sg, ding0093@e.ntu.edu.sg).
\IEEEcompsocthanksitem
Lvlong Lai and Qingyao Wu are with School of Software Engineering, South China University of Technology, and Pazhou Lab, Guangzhou, China (email: selailvlong@mail.scut.edu.cn, qyw@scut.edu.cn)

}

}

\maketitle

\input{0_abstract}

% \fi
\IEEEpeerreviewmaketitle

\input{1_introduction}

\input{2_related_work}

\input{3_method}

\input{4_experiment}

\input{5_conclusion}

\section*{Acknowledgment}
This research is supported by the National Research Foundation, Singapore under its AI Singapore Programme (AISG Award No: AISG-RP-2018-003), and the MOE Tier-1 research grants: RG28/18 (S), RG22/19 (S) and RG95/20.

\ifCLASSOPTIONcaptionsoff
  \newpage
\fi



% http://www.michaelshell.org/tex/bibtex/
\bibliographystyle{IEEEtran}
\bibliography{egbib}


% \end{thebibliography}

\begin{IEEEbiography}[{\includegraphics[width=1in,height=1.25in,clip]{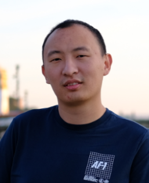}}]{Chi Zhang} is a PhD candidate with the School of Computer Science and Engineering, Nanyang Technological University, Singapore. He received the B.S. degree from China University of Mining and Technology in 2017.
His research interests are in computer vision and machine learning.
\end{IEEEbiography}

\begin{IEEEbiography}[{\includegraphics[width=1in,height=1.25in,clip]{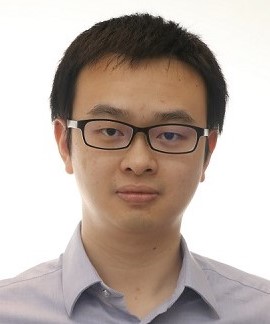}}]{Guosheng Lin} is an Assistant Professor at School of Computer Science and Engineering, Nanyang Technological University, Singapore. His research interests are in computer vision and machine learning.
\end{IEEEbiography}

\begin{IEEEbiography}[{\includegraphics[width=1in,height=1.25in,clip]{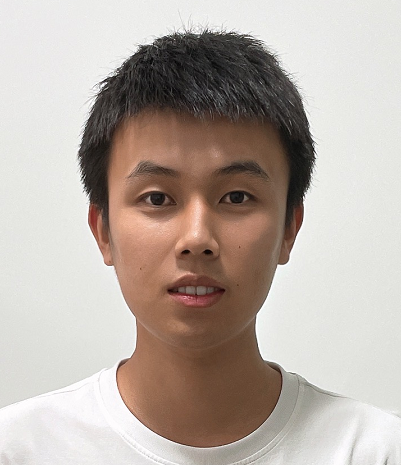}}]{Lvlong Lai}  is currently a Ph.D. student with the School of Software Engineering, South China University of Technology. His research interests include machine learning and computer vision.
\end{IEEEbiography}

\begin{IEEEbiography}[{\includegraphics[width=1in,height=1.25in,clip,keepaspectratio]{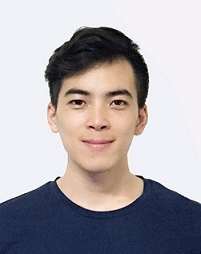}}]{Henghui Ding} received the B.E. degree from Xi'an Jiaotong University, Xi'an, China, in 2016. He received the Ph.D. degree from
Nanyang Technological University (NTU), Singapore, in 2020.
His research interests include computer vision and
machine learning.
\end{IEEEbiography}

\begin{IEEEbiography}[{\includegraphics[width=1in,height=1.25in,clip]{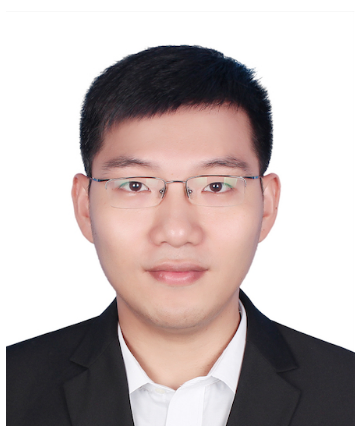}}]{Qingyao Wu} 
(Senior Member, IEEE) received the B.S. degree in software engineering from the South China University of Technology, China, in 2007, and the Ph.D. degree in computer science from the Harbin Institute of Technology, China, in 2013. He is currently a Professor with the School of Software Engineering, South China University of Technology. His current research interests include computer vision and data mining.
\end{IEEEbiography}


% \end{IEEEbiography}


% % insert where needed to balance the two columns on the last page with
% %\newpage



%\vfill


\end{document}

%% file: 0_abstract.tex
\begin{abstract}
Real-world visual recognition problems often exhibit long-tailed distributions, where the amount of data for learning in different categories shows  significant imbalance. 
Standard classification models learned on such data distribution often make biased predictions towards the head classes while generalizing poorly to the tail classes.
In this paper, we present two  effective modifications of CNNs to improve network learning from long-tailed  distribution.
First, we present a Class Activation Map Calibration (CAMC) module to improve  the learning and prediction of network classifiers, by enforcing network prediction based on important image regions.
The proposed CAMC module  highlights the correlated image regions across data and reinforces the representations in these areas to obtain a better global representation for classification.
Furthermore,  we investigate the use of normalized classifiers for representation learning in long-tailed problems. Our empirical study demonstrates that by simply scaling the outputs of the classifier with an appropriate scalar, we can effectively improve the classification accuracy on tail classes without losing the accuracy of head classes. 
We conduct extensive experiments to validate the effectiveness of our design and we set new state-of-the-art performance on five benchmarks, including ImageNet-LT, Places-LT, iNaturalist 2018, CIFAR10-LT, and CIFAR100-LT.

\end{abstract}

%% file: 1_introduction.tex
\section{Introduction}

\IEEEPARstart{O}{ver} the past decade, Deep Neural Networks have shown remarkable success in a broad range of computer vision tasks~\cite{liu2020weakly,zhang2021cyclesegnet,sun2020conditional,zhang2018efficient,sun2020open,sun2021m2iosr,Zhang_2021_CVPR,zhang2021navigator,liu2021fewshot}. A crucial reason is the availability of large-scale datasets, such as ImageNet~\cite{imagenet} to enable the understanding of visual concepts with high variance.
However, these datasets are often intentionally balanced, while real-world visual recognition exhibit long-tailed distributions, where the amount of data for each category may highly vary. 
Due to the data-hungry nature of deep networks, models learned from such unbalanced data distribution are often biased towards the head classes with sufficient training data and generalize poorly to tail classes.

\begin{figure}[t]
	\centering
	\includegraphics[width=1\linewidth]{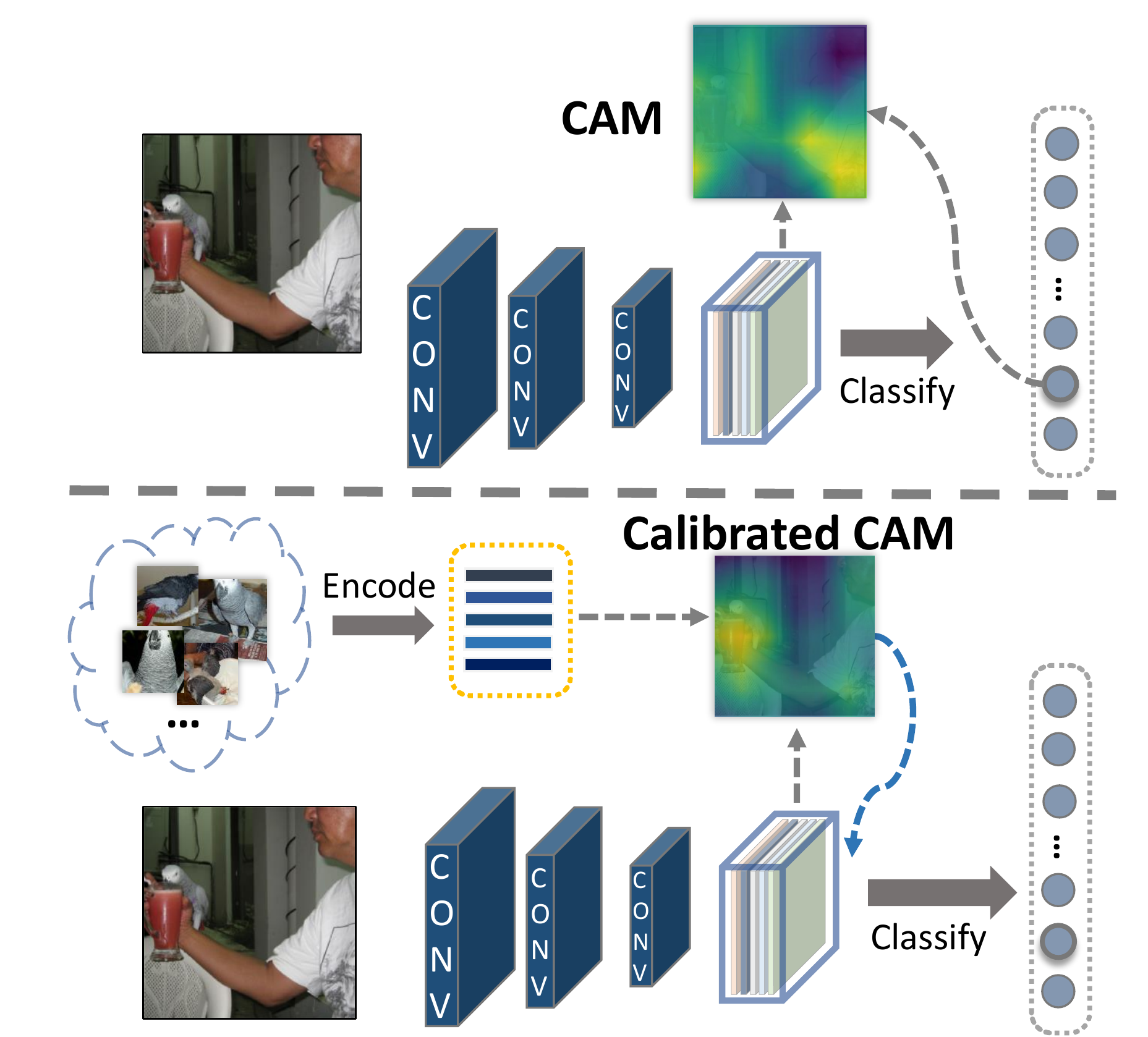}
	
	\caption{ The vanilla class activation mapping generates skewed CAMs for the tail class `grey bird' (up). We explicitly calibrate the CAMs which are used to generate better data representations for classification (bottom).
	}
	\label{fig:introduction}
\end{figure}

Many early studies solve data imbalance problems by artificially balancing the training, such as balanced data sampling strategies~\cite{over_sampling, over_sampling_2, under_sampling, class_balancing_1} and class-sensitive loss functions~\cite{ldam,distribution_balanced_loss, equalization,focal_loss,meta_softmax}, in the hope of assigning equal learning opportunities for each class.
However, it is commonly evidenced that although these balancing heuristics can alleviate the data imbalance problem to some extent, they often come with the price of significantly degrading performance of head classes and the potential risk of overfitting, 
and many of these methods may completely fail on the large-scale long-tailed datasets which exhibit severe data imbalance.

\begin{figure}[t]
\centering
	\includegraphics[width=0.9\linewidth]{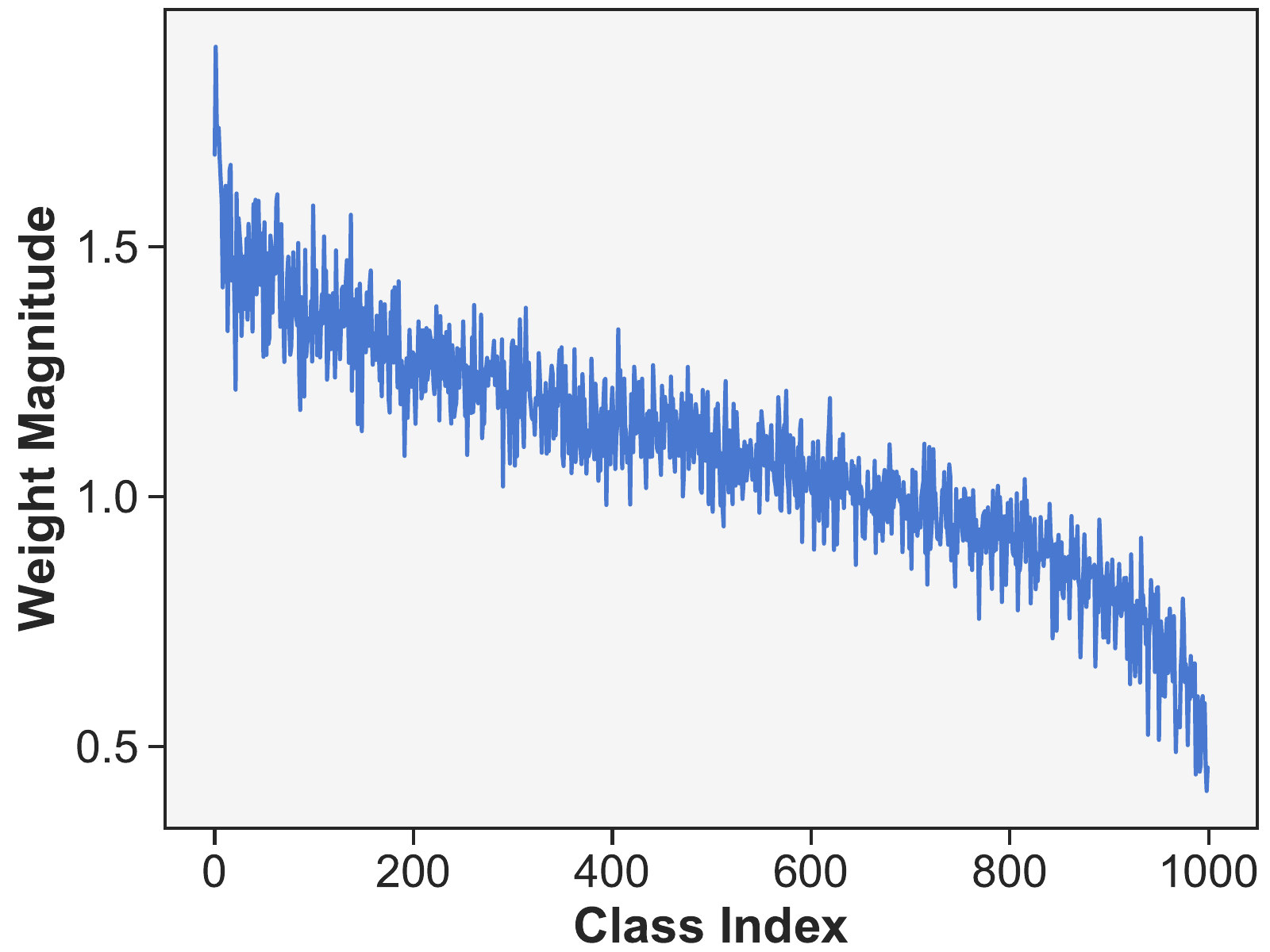}
	\caption{Illustration of the bias issue in a learned classifier after representation learning. 1000 classes in the ImageNet-LT dataset are sorted based on the number of training images and we plot the weight magnitude corresponding to each class. The magnitude of  weight vectors is positively correlated to the number of training images, which results in a bias issue in classification decisions.}
	\label{fig:bias}
\end{figure}

More recently, the studies in \cite{decoupling} and \cite{bbn} demonstrate that the classifier in the network particularly suffers from such unbalanced data distribution.
Concretely, the magnitudes of weights corresponding to each class are positively correlated with the number of data in a learned model, as shown in Fig.~\ref{fig:bias}. 
As a consequence, the network always produces biased logits toward the head classes. 
As is observed in~\cite{decoupling}, by merely re-training the classifier of a naturally trained CNN with a class-balanced data sampling strategy, the bias issue can be largely alleviated with remarkable performance improvement.

Nevertheless, such decoupled learning strategies will still not suffice to generate good decision boundaries to classify all classes.
During network training, the classifier is optimized to generate a good decision boundary by extracting shared and discriminative  information from each class. However, as the data from tail classes are very limited, the implicitly learned classifier may overlook those instance-specific semantics of tail classes, due to the high variations in limited data points, which may result in a waste of information in these categories. Moreover, the extracted discriminative semantics may also be highly noisy and inaccurate.
Such problems can be observed by visualizing the Class Activation Maps~\cite{cam}  (CAMs) of a learned classifier in Fig.~\ref{fig:introduction} (up) and Fig.~\ref{fig:CAMs}. As can be seen, the 
 activation regions of the tail classes often attend to the irrelevant areas in the images, which indicates that the learned classifier fail to  depict the  true distribution of tail classes and generates bad decision boundaries.
This suggests the need to differentiate the high-variance object regions from the cluttered background, as a crucial step to learn good decision boundaries.

To alleviate these problems, we present a CAM Calibration (CAMC) module ahead of the fully connected layers to force the network to pay more attention to the important regions in the image based on the correlations between images. 
Our design takes inspiration from the relation-based few-shot learning literature~\cite{SnellSZ17,zhang2020deepemd,zhang2020deepemdv2,zhang2019canet,pgnet} where the model directly makes predictions based on explicit data relations, which bypassing a difficult implicit classifier learning process in the low-shot case where information may be lost.
In our work, we incorporate the idea of using explicit data correlations into implicit classifier learning to calibrate the CAMs for better classifier learning and prediction.
The CAMC module caches a collection of  prototype vectors \emph{for each tail class}, which are initialized by the data embeddings.
By convolving the original feature maps with these prototypes, we can obtain a group of activation maps that highlight the correlated semantics shared across data, which are more likely to relate to the target classes.
We then reinforce the feature representations in the regions highlighted by the CAMC to force the classifier to have a larger response with these important regions, which hence improves the classifier learning and prediction.
Based on the explicit intervention in CAMC, the classifier always makes predictions  based more on the highlighted areas, which corrects the skewed attentions in the classifier.
The proposed CAMC can be inserted as a plug-and-play module in many classification networks. Despite its simplicity, the proposed CAMC module can effectively improve the network prediction. An illustration of our motivation is shown in Fig.~\ref{fig:introduction} (bellow).

The second problem is the current decoupled training pipeline is that 
at the representation learning stage, as the learning of the backbone relies on the gradients back-propagated from the classifier,
a biased classifier will inevitably have a negative impact on the learned representations.
As a consequence, the minimization of empirical error is achieved not only by learning discriminative representations in the backbones but also by enlarging the bias in the classifier.
Previous works~\cite{decoupling} have discussed normalized classifiers during classifier re-training, while 
we argue that it is also helpful to remove the classifier bias at the representation learning stage to facilitate the optimization of the network backbone.
However, we observe that directly normalizing the classifier weights, as was done in ~\cite{decoupling}, does not yield better performance over a standard linear classifier, while by simply adjusting the entropy of output  with a scaling scalar, we can effectively improve the performance of a CNN model for long-tailed recognition, and the normalization strategy can  benefits both the representation learning and the classifier decision.
We empirically show that the choice of the scalar values has a large influence on the generalization ability of the learned model, in particular for tail classes. 
Our normalization strategy is simple and easy to implement and does not requite the knowledge of class distributions $p(c)$, which is different with the  class-specific balancing strategy in other works~\cite{menon2020long,under_sampling_2,under_sampling,over_sampling_2,over_sampling}.

To sum up, our work makes improvements over previous works from the representation learning and classifier learning aspects.
To validate the effectiveness of our algorithm, we conduct comprehensive experiments on multiple datasets.
The contributions of this paper are summarized as follows:
\begin{itemize}
	\item We investigate the use of normalized classifiers for representation learning in long-tailed recognition.
	\item We propose a CAM calibration module to improve the learning and prediction of the classifier.
    \item Experiments on five popular benchmark datasets, including ImageNet-LT, Places-LT, iNaturalist 2018, CIFAR10-LT, and CIFAR100-LT, show that our method significantly outperforms the baselines and sets new state-of-the-art results.

\end{itemize}

%% file: 2_related_work.tex
\section{Related Work}
In this section, we review previous literature on the long-tailed recognition task and  then discuss other related topics to our paper.
\begin{figure*}[t]
	\centering
	\includegraphics[width=1\linewidth]{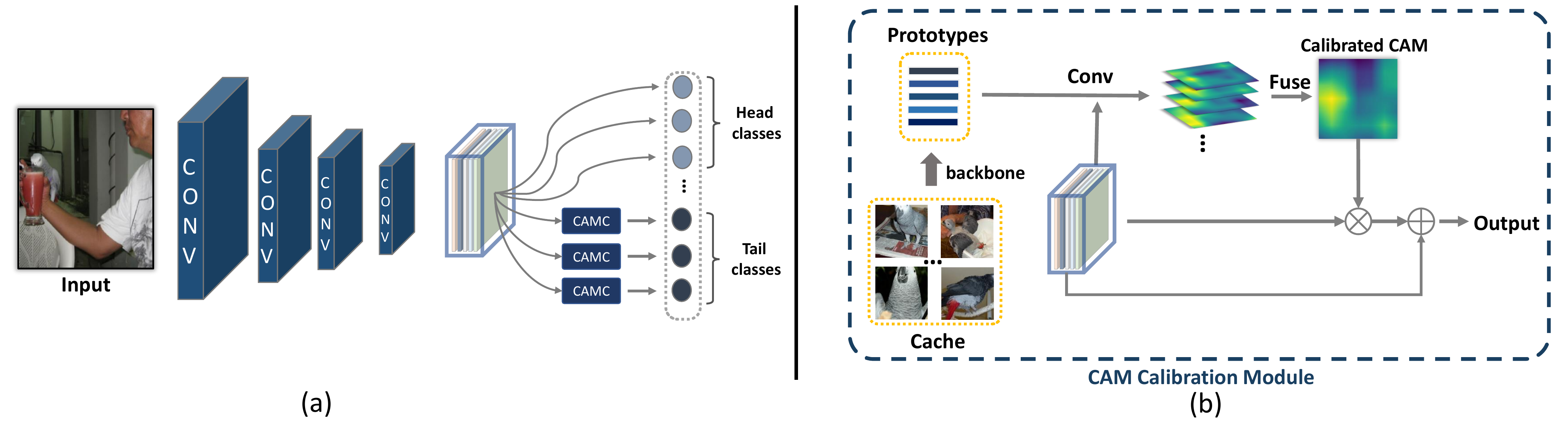}
	\caption{(a) Our framework for long-tailed recognition. (b) The proposed CAM calibration module (CAMC). In CAMC, cached data from each tail class are used as prototypes to generate the CAMs that highlight the correlated regions with these prototypes. We reinforce the representations in these regions to obtain an improved representation  to compute the scores for each tail class. (The global average pooling operator and the classifier are not indicated in this figure.)
	}
	\label{fig:main}
\end{figure*}

\textbf{Long-tailed recognition.} 
Research literature on long-tailed recognition exhibits great diversity.
The most dominant direction in handling data imbalance is to balance the training, in order to  obtain a more balanced data distribution. A line of efforts proposes to make modifications to the data sampling strategies, such as over-sampling data  from the minority classes\cite{over_sampling, over_sampling_2}, under-sampling data from majority classes~\cite{under_sampling, under_sampling_2}, or class-balanced sampling methods based on the number of data in the classes\cite{class_balancing_1}. 
\cite{ldam,distribution_balanced_loss, equalization, striking, hard_mining,focal_loss,meta_softmax, seesaw, balanced-activation}. For example, large weights are assigned to the training samples from the tail classes.
Cao~\etal~\cite{ldam} design a distribution-aware loss to enlarge the  margins for minority classes.
Apart from the aforementioned direction, recent works on long-tailed recognition also include methods based on transfer learning~\cite{transfer_1, transfer_2, transfer_3,transfer_4, tf_multi_experts, mitigating}, meta learning~\cite{oltr,l2rw,rethinking_methods,meta_weight_net}, metric learning~\cite{range_loss}, mixup\cite{mixup,manifold_mixup,remix}, self-supervised learning~\cite{rethinking_labels}, etc~\cite{etc_factors,inflated,etc_robust,etc_face,etc_sampling,etc_fea_aug,etc_keep_good, devil, RTC}.

Recent studies  in \cite{decoupling} and \cite{bbn} show that although balanced training can improve the performance of tail classes, they may harm representation learning. To handle this issue, Kang~\etal~\cite{decoupling}  proposes a simple decoupled training strategy that  re-trains the classifier with the class-balanced sampling strategy upon a fixed feature encoder that is naturally trained.
Our method in this paper also follows such a two-stage learning pipeline.

\textbf{ Normalized classifiers.} Normalized classifiers are shown to be useful in many visual recognition problems. Chen~\etal~\cite{closer_look} present a powerful baseline for few-shot learning by simply finetuning a cosine classifier upon a pre-trained data encoder. Hou~\etal~\cite{increment} adopt a cosine classifier for incremental learning which explicitly removes the bias in the magnitude between old and new classes.
Cao~\etal~\cite{ldam} adopt a cosine classifier to better tune the class margin for long-tailed classification.
Recently, Kang~\etal~\cite{decoupling} discuss the use of the  normalized classifiers and weight normalization~\cite{weight_normalization} for  classifier learning  in the long-tailed recognition problems.
However, they only apply normalization at the classifier re-training stage based on a naturally learned backbone, while our work focuses on the use of normalized classifiers for representation learning. 
We also emphasize the importance of the scaling factor to the long-tailed classification task and demonstrates that directly normalizing the classifier weights without scaling the entropy can not yield improved results over the vanilla classifier.

\textbf{Class Activation Maps.} Class Activation Mapping~\cite{cam} is able  to locate  discriminative parts of the objects from different classes based on the classifier weights. Since it can effectively establish meaningful correlations between image regions and class labels, it has been widely used for  weakly supervised learning with image labels.
Many variants of CAM  have also been proposed in the literature to improve the original activated regions, such as region erasing and expanding~\cite{erase_1}.
With a similar goal, model explainability also seeks to locate the regions corresponding to the network neurons. Previous works in this domain mainly include methods based on gradients~\cite{cam_grad_1, cam_grad_2} and  counterfactual reasoning~\cite{cam_counter_1, cam_counter_2}. 
Due to CAM's simplicity  and training-free properties, it is widely been used as a tool to evaluate the quality of a learned classifier.

%% file: 3_method.tex
\section{Method}
In this section, we present our framework for the long-tail recognition task.
We first have a brief review of the vanilla formulation in the Class Activation Mapping in Section ~\ref{sec:cam}.
Then we present our proposed CAM calibration module in Section~\ref{sec:camc}.
Finally, we describe the normalized classifiers adopted in our network for representation learning  in Section~\ref{sec:normalized}. 
Similar to ~\cite{decoupling}, our network also has two learning stages, including a representation learning stage and a classifier learning stage. \emph{Our proposed CAM calibration module is only applied at the second learning stage.}
The network structure of our model is illustrated in Fig.~\ref{fig:main}.

\subsection{Revisiting Class Activation Mapping}
\label{sec:cam}
Class Activation Mapping~\cite{cam} aims to locate the important image region that contributes to the decision behavior in a trained model. It elegantly turns a classifier into a class activation map detector without any additional training efforts.
Specifically, let $ \mathbf{F} \in \mathbb{R}^{H\times W\times C}$ denote the feature maps generated by the convolutional layers before the global average pooling operator, and 
$f_i(x,y)$ denotes the activation value at the spatial location (x,y) of the $i$th channel in $ \mathbf{F}$.
$\mathbf{W} \in \mathbb{R}^{N\times C} $ is the weight matrix in the classifier where each row vector $\mathbf{w}^{c}$ in $\mathbf{W}$  corresponds to a specific class $c$. $N$ is the number of classes and $C$ is the number of the feature dimensions.
The class activation map $\mathbf{M}_c(x,y)$ for class $c$ is defined as the weighted sum of the response maps in different channels of $F$ based on the class weight $\mathbf{w}^c$:
\begin{equation}
    \mathbf{M}_c(x,y)=\sum_{i}^{}\mathbf{w}^{c}_{i}f_i(x,y),
\end{equation}
where $\mathbf{w}^{c}_{i}$ is the value in the  $i$th dimension of $\mathbf{w}^{c}$.
Intuitively, the values in $\mathbf{w}^c$ essentially indicate the importance of different  channels to the prediction of class $c$. Therefore, by computing a weighted sum of the feature  response maps of all channels, we emphasize the region that has high response values at the important channels and these regions are deemed to make more contributions to the classification of the class.

\begin{figure}[t]
	\centering
	\includegraphics[width=0.9\linewidth]{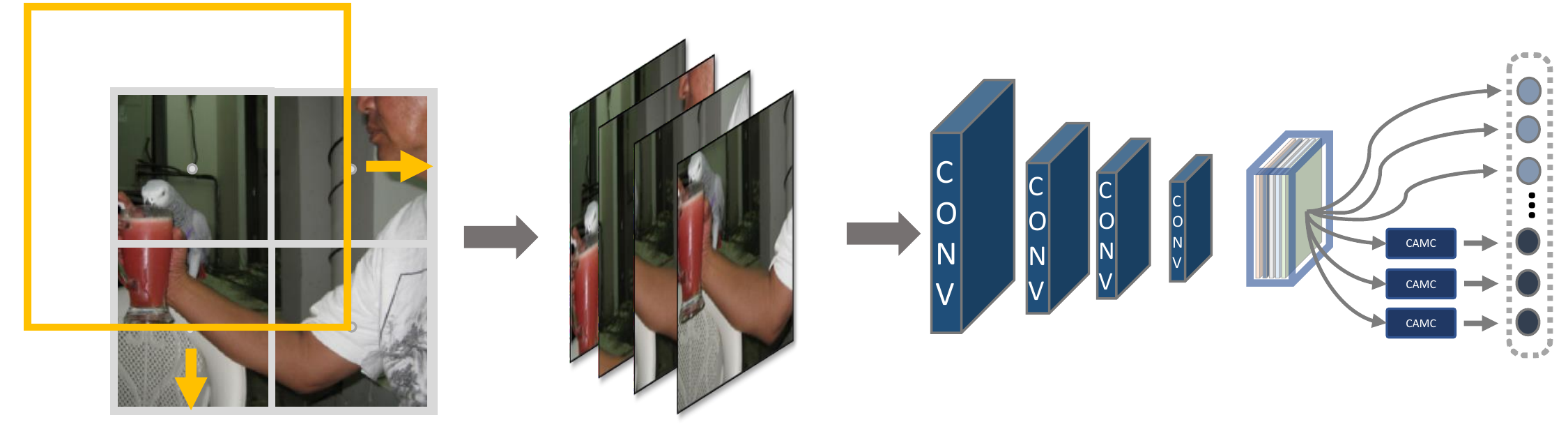}
	\caption{Illustration of CAMC++. We use a sliding window to crop the image into patches for processing. After obtaining the feature maps, we can reuse the operations in CAMC to make predictions.
	}
	\label{fig:camc_pp}
	%\vskip -1.5em
\end{figure}

An alternative interpretation of the class activation mapping is that the classifier learns  a prototype vector  $\mathbf{w}^c$  for each class, and by computing the inner product between the prototypes and the feature vector at each location, we can obtain a response map $ \mathbf{M}_c$ that highlights the  locations that show high similarity with the prototype  $\mathbf{w}^c$:
%\vskip -0.5em
\begin{equation}
\label{eq:sim_proto}
    \mathbf{M}_c(x,y)= \langle\mathbf{w}_c,f(x,y)\rangle,
\end{equation}
where $\langle\cdot,\cdot\rangle$ denotes the dot product operator and $f(x,y)$ is the feature vector at location $(x,y)$.
This also amounts to  transform the original CNN into a fully convolutional manner that
 classifies each location in the feature map densely with the classifier, \ie, 
 %\vskip -0.5em
 \begin{equation}
 \label{eq:conv}
    \mathbf{M}= \mathbf{F}* {\mathbf{W}},
\end{equation}
where $*$ denotes the 2D convolution operator and ${\mathbf{W}}$ is reshaped to $N\times C\times 1\times 1$ as the $1\times 1$ convolutional kernel. We omit the bias term in the above equations due to its negligible influence on the final predictions.
Therefore, the CAMs can also be seen as the score map of a dense prediction model.
As the derivation of CAMs is purely training-free and easy-to-implement, it is  widely used as a tool to qualitatively evaluate a learned CNN model.

\subsection{Calibrating CAMs for Long-Tailed Recognition}
\label{sec:camc}
For the long-tailed  image classification task, since the amount of data of tail classes is very limited, it raises  difficulty for the network to sufficiently understand these classes and generate the decision boundaries.
By visualizing the CAMs generated from a CNN  trained on the long-tailed dataset in Fig.~\ref{fig:introduction}, we find that the learned classifier often attends to irrelevant image regions for tail classes, 
which indicates that the implicitly learned classifier may confuse the background  and the target objects, and thus the learned classifier fails to accurately capture the real distribution of tail classes.

To mitigate this problem,  we design a CAM Calibration (CAMC) module to improve the learning and prediction of the classifier by forcing the classifier weights to have larger responses with the important regions in the images.
Our design is based on the intuition that as the qualities of the CAMs and the classifier are closely correlated, we can improve the quality of the classifier by improving the CAMs.
In order to locate the important region in the images, we assume that the semantics that are co-occurrent  across data in a category are more likely to belong to the target objects and should be assigned with more importance. By locating the shared semantics across data, we can not only better locate  the target object for prediction, but also regularize the classifier training. 
To this end, we let the data embeddings from this class play the role of the prototypes in Eq.\ref{eq:sim_proto}, such that each prototype  generates a CAM  that highlights the region correlated to it. 
Specifically,
Let  $\bf{\Phi}^{c} \in \mathbb{R}^{C \times K}$
be a collection of learnable prototype vectors initialized by  embedding vectors $[{\bf{\phi}}^{c}_1,...,{\bf{\phi}}^{c}_K]$  that are generated by $K$  images from class $c$. \emph{We use the backbone trained at the representation stage to encode  these data to generate prototypes.} We then use these prototypes as convolutional filters to convolve the  the feature map $\mathbf{F}$, as done in Eq.~\ref{eq:conv} , which results in a group of response maps with $K$ channels:
%\vskip -0.5em
 \begin{equation}
    \tilde{\mathbf{M}}_c= \mathbf{F} * \bf{\Phi}^{c},
\end{equation}
where $ \tilde{\mathbf{M}}_c \in \mathbb{R}^{H\times W\times K}$is the resulting  
 maps, and $\bf{\Phi}^{c}$ is reshaped to $K\times C\times 1\times 1$ as convolution kernels.
Next, different maps are fused by convolutions to generate a single-channel map $\hat{\mathbf{M}}_c \in \mathbb{R}^{H\times W\times 1}$.
Based on the new CAM, we reinforce the representation in the highlighted regions  before the  global average pooling layer in a residual manner:
%\vskip -0.5em
 \begin{equation}
    \tilde{\mathbf{F}}_c = (1+\sigma(\hat{\mathbf{M}}_c ))\circ \mathbf{F}_c,
\end{equation}
where  $\sigma(\cdot)$ is the sigmoid function and $\circ$ denotes Hadamard product.
Finally, we apply global average pooing to the refined feature representation to obtain a global representation $\mathbf{x}_c \in \mathbb{R}^{C}$, and generate the class score $s_c$ with the network classifier:
%\vskip -0.5em
\begin{equation}
\label{eq:score}
    \mathbf{s}_c=\langle  \frac{\mathbf{x}_c}{|\mathbf{x}_c|}, \mathbf{w}_c\rangle.
\end{equation}
We repeat such operations to generate the embeddings  for all the tail classes where the number of training images is less than a threshold $\tau$ .
For  the  embeddings of other classes, we simply  skip the proposed CAMC module and directly use the original global representations.
Since the scores of each class are generated based on their own modified version of the data embeddings, \ie, $\mathbf{x}_i$, we normalize the representations when computing the score in Eq.~\ref{eq:score}  to avoid the bias issue in the feature magnitude.

\textbf{CAMC++.} In addition  to the aforementioned CAMC module, we also present a design variant, denoted by CAMC++.
With the same purpose of using the rectified CAMs to re-weight feature points for obtaining a refined global representation, CAMC++  obtains the dense representations by cropping the input image into many patches before it is fed to the network. This can better encode local features based on patches without being influenced by irrelevant contexts.
As is shown in Fig.~\ref{fig:camc_pp}, we first use a sliding window with the raw image size to obtain $M\times M$ cropped images that are centered at each grid location and then resize these patches to the same input size.
Then, each cropped image is encoded by \emph{the convolutional layers and the global average pooling layer} to generate a  vector representation.  Therefore, the $M\times M$  vectors together construct a feature map $\mathbf{F \in \mathbb{R}^{M\times M\times C}}$, and we can re-use the operations in the CAMC module to generate the scores for classification.
The reason we make the sliding windows have the same size as the original image size is to avoid the discrepancy in the image scale  between two training stages, such that we can directly use the trained backbone in the first stage to encode patches. When the sliding window attends to regions out of the images, we only keep the image region without padding.

\subsection{Representation Learning with Normalized Classifiers}
\label{sec:normalized}
Before the aforementioned  classifier learning, we need to pre-train a  network backbone to learn representations for image encoding.
In a standard learning paradigm, a  CNN is trained to minimize its average error over the training data, which is known as the Empirical Risk Minimization principle~\cite{ERMP}.
As the empirical data distribution for network learning is significantly unbalanced in long-tailed recognition, the learning of the model is inevitably biased  to favor the prediction of the majority classes.
As is recently observed in ~\cite{decoupling}, a conspicuous bias issue in long-tailed recognition is that the magnitudes of weights in the linear classifier are positively correlated to the number of data points in the corresponding classes. 
This indicates that the empirical error minimization over the training data distribution is achieved not only by learning discriminative representations but also by enlarging the bias in the classification decision during the network training.
It is therefore important to remove the distraction of the classifier bias in the representation learning stage for learning discriminative representations.
To this end, we adopt a normalized classifier that simply assigns a fixed magnitude $g$ to scale the weight $\mathbf{w}_i$ corresponding to each class, and the score of a specific class $c$ in the first training stage is computed by:
%\vskip -0.5em
\begin{equation}
    \mathbf{s}_c=\langle  \mathbf{x}, g\frac{\mathbf{w}_c}{|\mathbf{w}_c|}\rangle.
\end{equation}
and in the  classifier-retraining stage, the score computation in the classifier learning stage (Eq.~\ref{eq:score}) becomes
%\vskip -0.5em
\begin{equation}
\label{eq:cosine}
    \mathbf{s}_c=\langle  \frac{\mathbf{x}_c}{|\mathbf{x}_c|}, g\frac{\mathbf{w}_c}{|\mathbf{w}_c|}\rangle.
\end{equation}
Obviously, a large scaling factor $g$ results in a low-entropy prediction after Softmax layer, while a small value increases the entropy.
In our experiment, we demonstrate that the choice of the scaling factor has a crucial influence on the convergence of the network training and the generalization ability of the model, particularly for the tail classes.
For example, in the classifier re-training stage, if we ignore $g$ in Eq.~\ref{eq:cosine}, the output scores will lie in $[-1,1]$, which results in very high entropy in the output probability, and the network training may completely fail.

\begin{figure*}[t]
	\centering
	\includegraphics[width=1\linewidth]{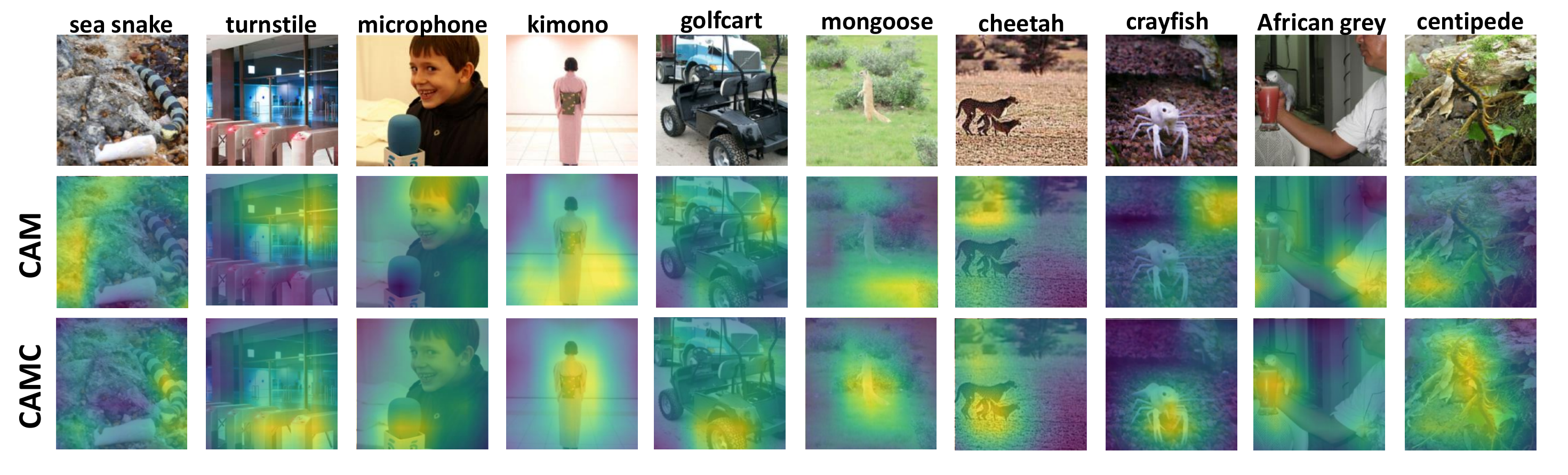}

	\caption{Comparison of the CAMs of tail classes generated by the vanilla class activation mapping~\cite{cam} and the  CAMC module. Our proposed CAMC effectively rectifies the activation regions for tail classes. 
	}
	\label{fig:CAMs}

\end{figure*}

\begin{figure*}[t]
	\centering
	\includegraphics[width=1\linewidth]{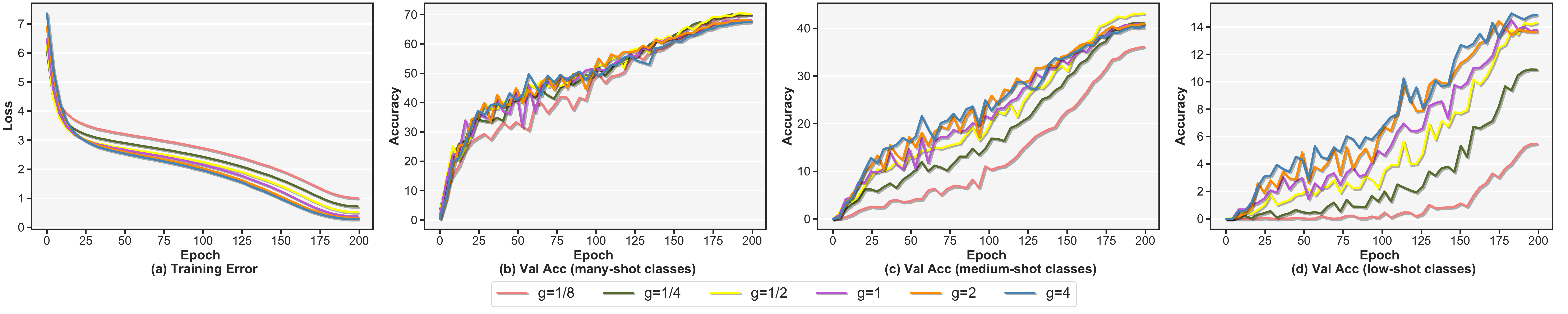}
	%\vskip -1em
	\caption{Training loss curve and validation accuracy curves under different magnitude values. Medium-shot and low-shot classes are  sensitive to the choice of magnitude values. }
	\label{fig:loss_acc}
\end{figure*}

\begin{figure}[t]
	\centering
	\includegraphics[width=1\linewidth]{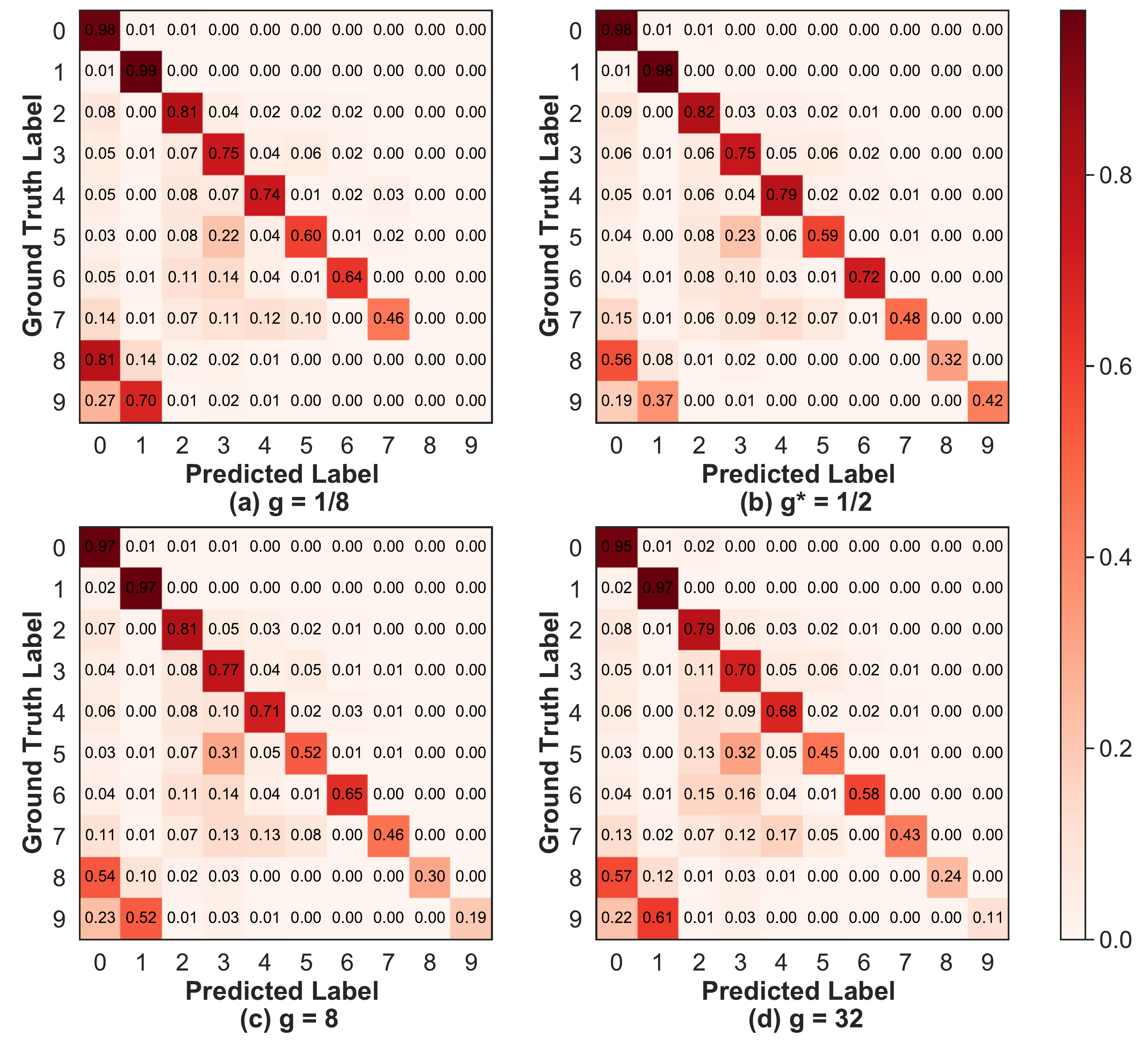}
	%\vskip -0.5em
	\caption{Confusion matrices under different magnitude values on the CIFAR10-LT-200 dataset, where the diagonal indicates the ground-truth.  Generalization performance of tail classes is sensitive to the magnitude values. Both very small or large values result in poor validation accuracy of tail classes. 
	}
	\label{fig:confusion_matrix}

\end{figure}

\begin{figure}[t]
	\centering
	\includegraphics[width=1\linewidth]{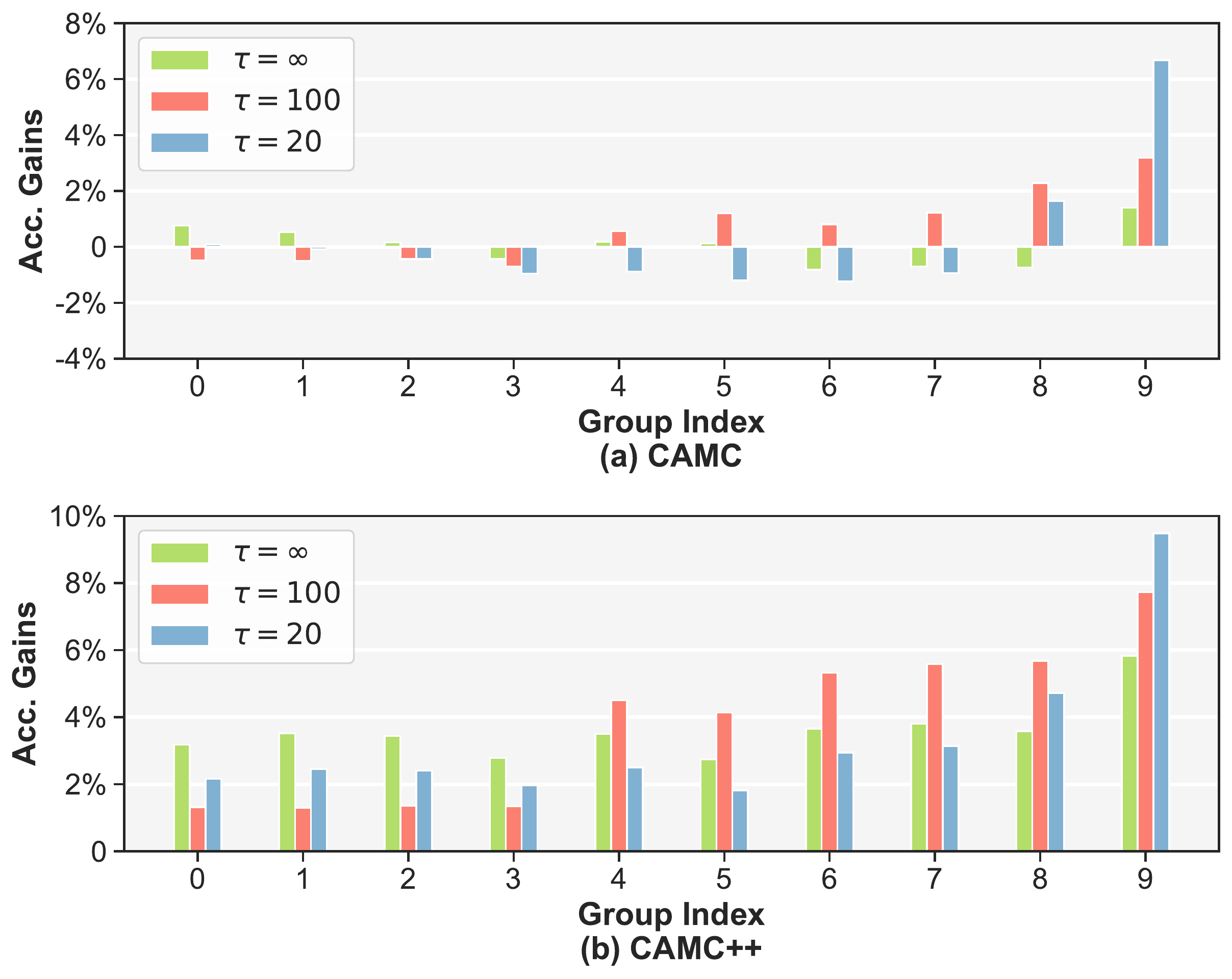}

	\caption{The performance gains achieved by CAMC (\textbf{a}) and CAMC++ (\textbf{b}) over the baseline under different threshold values. 1000 classes in the ImageNet-LT dataset are sorted and divided into 10 groups.  Applying CAMC and CAMC++ to both medium-shot classes and low-shot classes obtains the optimal balance between performance gains and losses on different groups. Low-shot classes have larger performance gains with our design.}
	\label{fig:cmcc_acc_gains}

\end{figure}

%% file: 4_experiment.tex
\section{Experiment}

\input{sub_content/0_cifar_dataset}
\input{sub_content/1_table_classifier}
\input{sub_content/2_camc}

\input{sub_content/3_sota1}
\input{sub_content/6_sota_cifar}

\subsection{Experiment Details.}

\textbf{Datasets.}
To validate effectiveness of our contributions, we conduct  experiments on 5 popular long-tailed recognition benchmarks, including ImageNet-LT\cite{oltr}, Places-LT~\cite{oltr}, iNaturalist 2018~\cite{inaturalist}, CIFAR10-LT~\cite{effective_numbers} and CIFAR100-LT~\cite{effective_numbers}.  We mainly report the results on the ImageNet-LT dataset for analysis and leave the results of other datasets in Section ~\ref{sec:sota}.
ImageNet-LT dataset is the truncated version of the ImageNet~\cite{imagenet} dataset, where the numbers of images in different classes range from 5 to 1280.  Following~\cite{decoupling}, we  divide the 1000 classes into many-shot classes (more than 100 images), medium-shot classes (20$\sim$100 images) and low-shot classes (less than 20 images) based on the number of images in the training set. We report the top-1 accuracy on all classes as well as on each split.

\textbf{Implementation}.
All the experiments in this paper are conducted with the PyTorch library~\cite{pytorch}. We mainly use ResNeXt-50~\cite{resneXt} as the backbone model to conduct experiments for analysis.
When comparing our model with previous works, we also report the performance of our model based on ResNeXt-152 and ResNet-\{10, 50, 152\}~\cite{resnet} backbones.
Most of our training hyper-parameters follows~\cite{decoupling}. If not specified, we use SGD optimizer with momentum of 0.9, batch size of 256, 200 training epochs for the  representation learning stage, 10 epochs for the CAMC training stage, and cosine learning rate scheduler  gradually decaying from 0.1 to 0.
The magnitude value $g$ is set as 0.5 and 16  in two learning stages, respectively, which are searched based on the validation set. 
We initialize 5 prototype vectors in the CAMC, as this is the  minimum number of training images in the classes of  ImageNet-LT.

\subsection{Analysis on Normalized Classifiers}
\label{exp:representation}
At the beginning, we investigate  the classifiers for representation learning. Apart from the normalized classifiers with an assigned magnitude (denoted by \textbf{Norm FC, \bm{$g^*$}} ), we also include the following classifiers for comparison:\\
\textbf{I. Linear.} This is the standard linear classifier in the CNNs.\\
\textbf{II. Weight Normalization}~\cite{weight_normalization}\textbf{.}Weight Normalization(\textbf{WN}) decouples the length of weight vectors from their direction, which is shown to be useful in standard image classification. This amounts to learning a class-wise magnitude $g_c$ for each class in the normalized classifier.\\
\textbf{III. Weight Normalization-\bm{$g$}}. A variant of weight normalization where the learnable magnitude of all classes are set as the same value. This is also equal to making the magnitude value  $g$ in our normalized classifier learnable. We initialize the magnitude with 1.\\
\textbf{IV. Norm FC, \bm{$g=1$}}. This is the L2 normalized classifier without an assigned magnitude value. Therefore, the weight vector corresponding to each class has a unit length, \ie, $g=1$.\\
To better evaluate the quality of the learned representations, we not only report 1) their performance under a standard CNN training paradigm, but also report the 2) performance when they are used as the fixed feature extractor to re-train a linear classifier with the class-balanced data sampling policy and 3)  their performance when they are used as the fixed feature extractor for the non-parametric  nearest class mean classifiers (\textbf{NCM}). The NCM parameterizes the classifier weights with the average training data in each class, and compute the cosine similarity for classification.
As we can see from the results in Table~\ref{table:classifier},
normalizing the classifier without assigning a magnitude value, as done in~\cite{decoupling}, can not boost  generalization performance over a standard linear classifier. 
After setting an appropriate magnitude value $g^*=0.5$, the normalized classifier can outperform the linear classifier under all three metrics by 1.7\%, 1.3\% and 1.4\%, respectively.
Note that unlike many other balancing strategies that come with the price of degrading the performance of head classes, \emph{the normalized classifier in our experiment can improve the performance on all three data splits}.
Learning a class-wise magnitude in \textbf{WN} can  slightly improve the performance of linear classifier. However, if we set the same learnable magnitude values for all classes, the performance drops significantly.

\textbf{Weight magnitude in the normalized classifier.}
We next investigate the influence of the magnitude value $g$ in the normalized  classifier.
We search the magnitude value $g$ exponentially from $2^{-5}$ to $2^{5}$ at the representation learning stage.
We plot the training loss curve and the validation accuracy curve of three splits in Fig.~\ref{fig:loss_acc}.
Here we choose the optimal value $g=1/2$ and some typical inappropriate values for illustration in the figure.
As we can see, the generalization performance of low-shot and medium-shot classes is particularly sensitive to the magnitude values. Both a very large value or a very  small value result in poor validation accuracy,
although a relatively larger magnitude value may result in a smaller empirical error on the biased training set.
To further observe the influence of the magnitude to different classes, we plot the confusion matrices of models trained with different magnitude values in Fig.~\ref{fig:confusion_matrix}. 
As we can see, the network with inappropriate magnitude values is more likely to confuse head and tail classes.
Specifically, if the magnitude is very small, the network outputs a high-entropy prediction and the optimization will penalize the cross-entropy loss more heavily. As the numbers of data between the confused head and tail classes (\eg class 0 and class 9 in the figure) are highly unbalanced, the network would simply ignore the data from tail classes and predict all confused data as head classes under such heavy penalization; likewise, if the magnitude is very large, the penalization to the loss becomes too weak before the confused tail classes are well classified.

\subsection{Analysis on CAMC}
 In Table~\ref{table:CAMC}, we compare the results of CAMC and CAMC++. We focus on the threshold value $\tau$ and the size of the crops in CAMC++.
In particular, when $\tau=\infty$, the CAMC module applies to all the classes and when $\tau=0$, the CAMC module is not used for any classes.
As the result shows, the optimal performance of CAMC is obtained when the threshold is set as 100, which means the CAMC is applied to the medium-shot and the low-shot classes. 
Applying the CAMC++ can improve the performance of all class sets, and performance grows consistently when the crop sizes increases from 2 to 5. 
To further analyze the influence of the CAMC module to different category, we sort all 
1000 categories in ImageNet-LT based on the number of training data and divide them into 10 groups. We observe the relative performance gains over the baseline.
As is shown in Fig.~\ref{fig:cmcc_acc_gains}, when the CAMC module applies to  all classes, the overall improvement is small, while only applying the CAMC to tail classes can significantly boost the performance of the tail classes but degrades the accuracy of medium-shot classes. When we apply CAMC to both medium-shot classes and low-shot classes, it achieves the optimal balance between performances gains and losses, and low-shot classes benefit more from the CAMC module.
Similar observations are also found for CAMC++, except that  there is no performance loss for all groups.

\textbf{Visualization of CAMs.}
To better understand the behavior in CAMC, we compare the CAMs generated by our model with the CAMs generated by the vanilla class activation mapping~\cite{cam} in Fig.~\ref{fig:CAMs}.
As we can see, the vanilla  CAMs generated by the classifier often attend  to irrelevant regions for data from tail classes, while our CAMC module  effectively rectifies the activation region in the images to make the classifier re-focus on the object area.

\subsection{Comparison with the State-of-the-Art}
\label{sec:sota}

 To better position our method among the long-tailed classification literature, we compare our performance with the state-of-the-art results on five benchmarks, namely, Imagenet-LT,  Places-LT, iNaturalist 2018,  CIFAR10-LT, and CIFAR100-LT  datasets.
\textbf{1) ImageNet-LT.} Apart from the ResNeXt-50 backbone, we also report our results with ResNet-10 and ResNeXt-152.
\textbf{2) Places-LT.} 
Places-LT is truncated from the Places2 dataset, which contains 365 classes and the number of images per class ranges from 5 to 4980.
Following ~\cite{oltr} and ~\cite{decoupling}, we report the performance with a ResNet-152 backbone pre-trained on ImageNet. 
\textbf{3) iNaturalist 2018.} 
iNaturalist 2018 is a large-scale long-tailed recognition dataset consisting data from 8,142 categories.
Following ~\cite{decoupling}, we report our model based on ResNet-50 backbone, which is  trained for 200 epochs.
\textbf{4) CIFAR10-LT and CIFAR100-LT.} CIFAR-10 and CIFAR-100 ~\cite{cifar} are balanced classification datasets which include 10 and 100 categories, respectively. There are  50,000 images for training and 10,000 images for testing in both datasets.
Following ~\cite{effective_numbers}, we use an imbalanced factor $\rho$ to truncate CIFAR-10 and CIFAR-100 and construct their long-tailed versions, CIFAR10-LT and CIFAR100-LT, respectively. $\rho$ is defined as $\max\{N_{i}\}/\min\{N_{i}\}$, where $N_{i}$ is number of samples of class $i$. Larger value of $\rho$ means the dataset suffers heavier long-tailed problem.
 Table \ref{table:cifar_dataset} shows the statistics of the dataset under different $\rho$ values.
The values of $\rho$ we used in the experiments include 10, 20, 50, 100 and 200. 
As the minimum number of data in the classes varies under different set-ups, we also adopt different thresholds $\tau$ for the proposed CAMC module. Specifically,  $\tau$ is 100 for all experiments in CIFAR100-LT dataset. For experiments in the CIFAR10-LT dataset, $\tau$ is 1000 when  $\rho$ is in $\{10, 20, 50\}$ and is 200 when   $\rho$ is in $\{100,200\}$.
The results on five benchmarks are shwon in Table~\ref{table:sota1} and Table~\ref{table:cifar_result}.
As can be seen, \emph{our improvements over previous works on different benchmarks are remarkable and consistent.} 
In particular, experiments on CIFAR-LT datasets show that our performance advantage grows when the imbalance ratio gets larger, which suggests our method is particularly useful for handling long-tailed issues.

%% file: sub_content/0_cifar_dataset.tex
\begin{table*}[t]
\centering

\small
\resizebox{0.85\textwidth}{!}{%
\begin{tabular}{l @{\extracolsep{\fill}} IcIcIcIcIcIcIcIcIcIc}
\Xhline{1pt}
\multicolumn{1}{cI}{\textbf{Dataset}} & \multicolumn{5}{cI}{\textbf{CIFAR10-LT}} & \multicolumn{5}{c}{\textbf{CIFAR100-LT}} \\
\Xhline{1pt}
\multicolumn{1}{cI}{\textbf{Imbalance Ratio} \bm{$\rho$}} & \multicolumn{1}{cI}{\textbf{10}} & \multicolumn{1}{cI}{\textbf{20}} & \multicolumn{1}{cI}{\textbf{50}} & \multicolumn{1}{cI}{\textbf{100}} & \multicolumn{1}{cI}{\textbf{200}} & \multicolumn{1}{cI}{\textbf{10}} & \multicolumn{1}{cI}{\textbf{20}} & \multicolumn{1}{cI}{\textbf{50}} & \multicolumn{1}{cI}{\textbf{100}} & \multicolumn{1}{c}{\textbf{200}} \\
\Xhline{1pt}
\textbf{Max. Number} & 5000 & 5000 & 5000 & 5000 & 5000 & 500 & 500 & 500 & 500 & 500 \\
\Xhline{1pt}
\textbf{Min. Number} & 500 & 250 & 100 & 50 & 25 & 50 & 25 & 10 & 5 & 2 \\
\Xhline{1pt}

\end{tabular}%
}
%\vskip -0.5em
\caption{
Statistics of CIFAR10-LT and CIFAR100-LT datasets. We present the maximum and minimum numbers of training images in the classes under different 
imbalance ratio $\rho$.
}
\label{table:cifar_dataset}
%\vskip -1em
\end{table*}

%% file: sub_content/1_table_classifier.tex
\begin{table*}[t]
    \small

% 	\begin{minipage}{\textwidth}
			%\vspace{20pt}
			\centering
			\resizebox{\textwidth}{!}{
				\begin{tabular*}{\linewidth}{l @{\extracolsep{\fill}}cccc IccccIcccc}%{lccccc} @{\extracolsep{\fill}} 
					\Xhline{1pt}
					\multicolumn{1}{c}{\multirow{2}{*}{\textbf{Classifier}}} & \multicolumn{4}{cI}{\textbf{Representation Learning}} & \multicolumn{4}{cI}{\textbf{Classifier Re-Training }}  & \multicolumn{4}{c}{\textbf{Nearest Class Mean Classifier}}                             \\ \Xcline{2-13}{1pt}%

					\multicolumn{1}{c}{}    & \multicolumn{1}{c}{\textbf{Many}} & \multicolumn{1}{c}{\textbf{Medium}} & \multicolumn{1}{c}{\textbf{Low}} & \multicolumn{1}{cI}{\textbf{All}} & \multicolumn{1}{c}{\textbf{Many}} & \multicolumn{1}{c}{\textbf{Medium}} & \multicolumn{1}{c}{\textbf{Low}} & \multicolumn{1}{cI}{\textbf{All}} & \multicolumn{1}{c}{\textbf{Many}} & \multicolumn{1}{c}{\textbf{Medium}} & \multicolumn{1}{c}{\textbf{Low}} & \multicolumn{1}{c}{\textbf{All}} \\%\midrule[0.5pt]
\Xhline{1pt}

\textbf{Linear (baseline)} & 66.4&	40.1&	12.0&	46.4 & 64.0 & 45.5 & 26.2 & 50.0 & 62.0 & 45.8 & 29.7 & 49.8 \\

\textbf{Weight Norm} & 66.9 &	40.2&	13.3	& 46.8 &64.6 & 45.9 & 27.1 & 50.5 & 63.2 & 45.3 & 29.3 & 50.0 \\

\textbf{Weight Norm-g} & 63.0&	35.4&	11.9&	42.8& 56.6 & 44.1 & 29.3 & 46.9  & 55.5 & 42.6 & 23.9 & 45.0\\

\textbf{Norm FC,} \bm{ $g=1$} & 66.5&	39.8&	14.0&	46.6 & 64.3 & 45.3 & 25.1 & 49.9 & 63.0 & 44.8 & 27.0 & 49.4 \\ 

\textbf{Norm FC,} \bm{ $g^*$} & \textbf{68.9}&\textbf{41.1}&\textbf{	14.2}&\textbf{	48.1} & \textbf{66.0} & \textbf{46.7} & \textbf{26.7} & \textbf{51.3} & \textbf{65.4} & \textbf{46.2} & \textbf{29.3} & \textbf{51.2} \\\Xhline{1pt}

\end{tabular*}
}
%\vskip -0.5em
\caption{ Evaluation of representations learned with different classifiers on the ImageNet-LT dataset with the ResNeXt-50 backbone. The normalized classifier with an appropriate  magnitude value (\textbf{Norm FC,} \bm{ $g^*$}) outperforms the linear classifier for representations learning  under three evaluation metrics consistently.} 
		\label{table:classifier}
		%\vskip -1.5em
\end{table*}

%% file: sub_content/2_camc.tex
\begin{table}[t]
\centering

\resizebox{0.45\textwidth}{!}{%
\begin{tabular}{lcIcccc}
\Xhline{1.2pt}
\multicolumn{1}{l}{\textbf{Method}}  & \bm{$\tau$}  & \textbf{Many} & \textbf{Medium}  &  \textbf{Low} & \textbf{All}\\ \Xhline{1pt}
\textbf{CAMC}& \textbf{0} & 66.0&	46.7&	26.7&	51.3         \\ 
\textbf{CAMC}& \textbf{20} & 67.2	&44.9&	32.1&	51.7        \\ 
\textbf{CAMC}& \textbf{100}  & 65.6&	48.0&	29.7&	\textbf{52.4}               \\
\textbf{CAMC}&  \bm{$\infty$} & 66.4&	46.6&	27.6&	51.6             \\
\Xhline{1pt}
\textbf{CAMC++,} \bm{$M=2$}&\textbf{100} &66.2&	50.1&	31.8&	\textbf{53.8}    \\ 
\textbf{CAMC++,} \bm{$M=3$}&\textbf{100} & 67.0&	51.5&	33.9&	\textbf{55.1}   \\ 
\textbf{CAMC++,} \bm{$M=4$}&\textbf{100} &67.4&	51.6&	34.3&	\textbf{55.3}    \\ 
\textbf{CAMC++,} \bm{ $M=5$} &\textbf{100}& 67.4&	51.8&	34.2&	\textbf{55.4}    \\ 
\Xhline{1pt}

\end{tabular}%
}
%\vskip -0.5em
\caption{Analysis on CAMC and CAMC++.  The optimal performance is achieved when CAMC is applied to medium-shot and low-shot classes, \ie, $\tau=100$.  The performance of CAMC++ grows consistently when the crop size $M$ increases from 2 to 5. }
\label{table:CAMC}
%\vskip -1.5em
\end{table}

%% file: sub_content/3_sota1.tex
\begin{table*}[t]
\centering
\small
\resizebox{0.9\textwidth}{!}{%
\begin{tabular}{lcccIcIc}
\Xhline{1pt}
\multicolumn{1}{c}{\multirow{2}{*}{\textbf{Methods}}} & \multicolumn{3}{cI}{\textbf{ImageNet\_LT}} & \multicolumn{1}{cI}{\textbf{Places\_LT}} & \multicolumn{1}{c}{\textbf{iNaturalist2018}} \\
\Xcline{2-6}{1pt}
\multicolumn{1}{c}{} & \multicolumn{1}{c}{\textbf{ResNet10}} & \multicolumn{1}{c}{\textbf{ResNeXt50}} & \multicolumn{1}{cI}{\textbf{ResNeXt152}} & \multicolumn{1}{cI}{\textbf{ResNet152}} & \multicolumn{1}{c}{\textbf{ResNet50}} \\
\Xhline{1pt}
\textbf{FSLwF}$\dagger$~\cite{fslwf} & 28.4 & - & - & 34.9 & - \\ 
\textbf{Focal Loss}$\dagger$~\cite{focal_loss} & 30.5 & - & - & 34.6 & -\\
\textbf{Range Loss}$\dagger$~\cite{range_loss} & 30.7 & - & - & 35.1 & - \\
\textbf{Lifted Loss}$\dagger$~\cite{lifted_loss} & 30.8 & - & - & 35.2 & - \\
\textbf{OLTR}$\dagger$~\cite{oltr} & 37.3 & 46.3 & 50.3 & 35.9 & - \\
\textbf{LDAM}~\cite{ldam} & - & - & - & - & 68.0 \\
\textbf{Effective Number}~\cite{effective_numbers} & - & - & - & - & 64.2 \\
\textbf{Rethink-DA}~\cite{rethinking_methods} & - & - & - & - & 67.6 \\
\textbf{Decoupling-NCM}~\cite{decoupling} & 35.5 & 47.3 & 51.3 & 36.4 & 63.1 \\
\textbf{Decoupling-cRT}~\cite{decoupling} & 41.8 & 49.5 & 52.4 & 36.7 & 67.6 \\
\textbf{Decoupling-}\bm{$\tau$}~\cite{decoupling} & 40.6 & 49.4 & 52.8 & 37.9 & 69.3 \\
\textbf{Decoupling-LWS}~\cite{decoupling} & 41.4 & 49.9 & 53.3 & 37.6 & 69.5 \\
\textbf{BBN}~\cite{bbn} & - & - & - & - & 69.6 \\
\textbf{IEM}~\cite{inflated} & 43.2 & - & - & 39.7 & 70.2 \\
\Xhline{1pt}
\textbf{CAMC} & \textbf{44.7} & \textbf{52.4} & \textbf{55.1}  &\textbf{39.1}   & \textbf{69.8}    \\
\textbf{CAMC++} & \textbf{47.2} & \textbf{55.4} &\textbf{58.5}  &\textbf{40.3}   & \textbf{73.0}    \\\Xhline{1pt}
\end{tabular}%
}
%\vskip -0.5em
\caption{Comparison of Top-1 Accuracy (\textbf{\%}) on ImageNet\_LT, Places\_LT and iNaturalist2018 datasets. $\dagger$ denotes results copied from~\cite{decoupling}. Our proposed method demonstrates remarkable performance advantages over  previous works on three benchmarks.} 
\label{table:sota1}
%\vskip -1.3em
\end{table*}

%% file: sub_content/6_sota_cifar.tex
\begin{table*}[t]
\centering

\small
\resizebox{0.9\textwidth}{!}{%
\begin{tabular}{l @{\extracolsep{\fill}} IcIcIcIcIcIcIcIcIcIc}
\Xhline{1pt}
\multicolumn{1}{cI}{\textbf{Dataset}} & \multicolumn{5}{cI}{\textbf{CIFAR10-LT}} & \multicolumn{5}{c}{\textbf{CIFAR100-LT}} \\
\Xhline{1pt}
\multicolumn{1}{cI}{\textbf{Imbalance Ratio}} & \multicolumn{1}{cI}{\textbf{10}} & \multicolumn{1}{cI}{\textbf{20}} & \multicolumn{1}{cI}{\textbf{50}} & \multicolumn{1}{cI}{\textbf{100}} & \multicolumn{1}{cI}{\textbf{200}} & \multicolumn{1}{cI}{\textbf{10}} & \multicolumn{1}{cI}{\textbf{20}} & \multicolumn{1}{cI}{\textbf{50}} & \multicolumn{1}{cI}{\textbf{100}} & \multicolumn{1}{c}{\textbf{200}} \\
\Xhline{1pt}
\textbf{Class-balanced Finetune}$\dagger$~\cite{cb_finetune} & 86.4 & 86.3 & 77.4  & 71.3    & 66.2 & 57.6 & 52.3 & 46.4 & 41.8 & 38.7 \\
\textbf{L2RW}$\dagger$~\cite{l2rw} & 85.2 & 83.1 & 78.9  & 74.2    & 66.5 & 53.7 & 51.6 & 44.4 & 40.2 & 33.4 \\
\textbf{Meta-Weight Net}$\dagger$~\cite{meta_weight_net} & 87.6 & 84.5 & 79.1  & 73.6    & 67.2 & 58.9 & 53.3 & 45.7 & 41.6 & 36.6 \\
\textbf{LDAM}~\cite{ldam} & 88.2 & - & -  & 77.0    & - & 58.7 & - & - & 42.0 & -     \\ 
\textbf{Effective Number}~\cite{effective_numbers} & 87.5 & 84.4 & 79.3  & 74.6    & 68.9 & 58.0 & 52.6 & 45.3 & 39.6 & 36.2     \\ 
\textbf{Rethink-DA}~\cite{rethinking_methods} & 88.9 & 86.5 & 80.5  & 76.4    & 70.7 & 59.6 & 55.6 & 48.5 & 43.4 & 39.3 \\
\textbf{BBN}~\cite{bbn} & 88.3 & - & 82.2  & 79.8    & - & 59.1 & - & 47.0 & 42.6 & -     \\
\Xhline{1pt} 
\textbf{CAMC} & \textbf{88.8} & \textbf{86.5} & \textbf{83.4}  & \textbf{79.4}   & \textbf{74.3} & \textbf{59.6} & \textbf{56.1} & \textbf{49.0} & \textbf{44.7} & \textbf{40.7}    \\ 
\textbf{CAMC++} & \textbf{88.0} & \textbf{85.3} & \textbf{81.7}  & \textbf{77.1}    & \textbf{72.9} & \textbf{57.2} & \textbf{54.2} & \textbf{46.8} & \textbf{43.3} & \textbf{38.6}        \\
\Xhline{1pt}

\end{tabular}%
}
%%\vskip -0.5em
\caption{Comparison of Top-1 Accuracy (\textbf{\%}) on CIFAR10-LT and CIFAR100-LT. $\dagger$ denotes results copied from~\cite{rethinking_methods}. Our proposed methods achieve better performance, particularly when the imbalance ratio is large.}
\label{table:cifar_result}
%%\vskip -1em
\end{table*}

%% file: 5_conclusion.tex
\section{Conclusion}
In this paper, we present two modifications of CNNs to address the long-tailed classification problems.
We first investigate the use of normalized classifiers for long-tailed visual recognition. By simply setting a normalization factor to the normalized classifier, we can effective improve both representation learning and classifier learning. We further present a CAM calibration module to enforce network prediction based on important regions in the image. Experiments on five benchmarks validate the effectiveness of our contributions, and we set new state-of-the-art performance on all of them.